\documentclass[runningheads]{llncs}
\usepackage{graphicx}
\usepackage{comment}
\usepackage{amsmath,amssymb} 
\usepackage{color}
\usepackage{xcolor}
\usepackage{graphicx}
\usepackage{float}
\usepackage{amsmath}
\usepackage{amsfonts}
\usepackage{amssymb}
\usepackage{multirow}
\usepackage{tabu}
\usepackage[title]{appendix}
\usepackage[english]{babel}
\usepackage{bm}
\usepackage{colortbl}
\usepackage{enumitem}
\usepackage{array}
\usepackage[ruled,vlined]{algorithm2e}
\usepackage{threeparttable}
\usepackage{dsfont}
\usepackage{pifont}
\usepackage{subcaption}
\usepackage{nccmath}
\usepackage{booktabs}
\usepackage{listings,lstautogobble}
\usepackage{dsfont}
\usepackage{pifont}
\usepackage{pbox}
\usepackage{tabularx}
\usepackage{graphicx}
\usepackage[super]{nth}

\usepackage{color, colortbl}
\definecolor{Gray}{gray}{0.9}
\definecolor{LightCyan}{rgb}{0.88,1,1}
\definecolor{carnationpink}{rgb}{1.0, 0.65, 0.79}

\newcolumntype{Y}{>{\centering\arraybackslash}X}

\newlength\savewidth\newcommand\shline{\noalign{\global\savewidth\arrayrulewidth
  \global\arrayrulewidth 1pt}\hline\noalign{\global\arrayrulewidth\savewidth}}
\newcolumntype{x}[1]{>{\centering\arraybackslash}p{#1pt}}

\newcolumntype{Y}{>{\centering\arraybackslash}X}

\graphicspath{{images/}}

\newcommand{\bnInplaceSync}{\textsc{InPlace-ABN$^{\mathsf{sync}}$}\xspace}

\begin{document}
\pagestyle{headings}
\mainmatter
\title{Segmentation Transformer:
Object-Contextual Representations for Semantic Segmentation\thanks{\color{darkgray}The OCR
(Object-Contextual Representation) approach is equivalent to the Transformer encoder-decode scheme.
We rephrase the OCR approach
using the Transformer language.}}

\titlerunning{Segmentation Transformer: OCR for Semantic Segmentation} 
\authorrunning{Yuhui Yuan, Xiaokang Chen\inst{3}, Xilin Chen, Jingdong Wang} 
\author{Yuhui Yuan\inst{1,2,3} \and
Xiaokang Chen\inst{3} \and
Xilin Chen\inst{1,2} \and
Jingdong Wang\inst{3}
}
\institute{
Key Lab of Intelligent Information Processing of Chinese Academy of Sciences (CAS), Institute of Computing Technology, CAS \and
University of Chinese Academy of Sciences \and
Microsoft Research Asia \\
\email{\{yuhui.yuan, v-xiaokc, jingdw\}@microsoft.com, xlchen@ict.ac.cn} }

\maketitle

\begin{abstract}
In this paper, we study the context aggregation problem in semantic segmentation.
Motivated by
that the label of a pixel is 
the category of the object that the pixel belongs to,
we present a simple yet effective approach,
object-contextual representations,
characterizing a pixel by 
exploiting the representation of the corresponding object class. 
First,
we learn object regions
under the supervision
of the ground-truth segmentation.
Second,
we compute the object region representation
by aggregating the representations
of the pixels lying in the object region.
Last,
we compute the relation between each pixel and each object region,
and augment the representation of each pixel
with the object-contextual representation
which is a weighted aggregation of all the object region representations.
We empirically demonstrate our method achieves competitive performance on various benchmarks:
Cityscapes, ADE20K, LIP, PASCAL-Context and COCO-Stuff.
Our submission ``HRNet + OCR + SegFix" achieves the \nth{1} place 
on the Cityscapes leaderboard by the ECCV 2020 submission deadline. 
Code is available at:
{\footnotesize\url{\color{blue}{https://git.io/openseg}}}
and {\footnotesize\url{\color{blue}{https://git.io/HRNet.OCR}}}.

~~~{\color{darkgray}We rephrase the object-contextual representation scheme
using the Transformer encoder-decoder framework.
The first two steps,
object region learning
and object region representation computation,
are integrated as 
the cross-attention module in the decoder:
the linear projections used
to classify the pixels, i.e., generate
the object regions,
are category queries,
and the object region representations
are the cross-attention outputs.
The last step
is the cross-attention module we add to the encoder,
where the keys and values are the decoder output
and the queries are
the representations at each position.
The details are presented in~Section~\ref{sec:segmentationtransformer}.
}

\end{abstract}

\section{Introduction}
\label{sec:introduction}

Semantic segmentation is a problem of assigning a class label
to each pixel for an image.
It is a fundamental topic in computer vision 
and is critical for various practical tasks 
such as autonomous driving.
Deep convolutional networks
since FCN~\cite{long2015fully}
have been the dominant solutions.
Various studies have been conducted,
including high-resolution representation learning~\cite{chen2018encoder,sun2019high},
contextual aggregation~\cite{zhao2017pyramid,chen2017rethinking}
that is the interest of this paper,
and so on. 

The context of one position
typically refers to 
a set of positions, e.g., the surrounding pixels.
The early study is mainly about the spatial scale of contexts,
i.e., the spatial scope.
Representative works, such as ASPP~\cite{chen2017rethinking} 
and PPM~\cite{zhao2017pyramid},
exploit multi-scale contexts.
Recently,
several works,
such as DANet~\cite{fu2018dual}, CFNet~\cite{zhang2019co} and OCNet~\cite{yuan2018ocnet,yuan2021},
consider
the relations between
a position and its contextual positions,
and aggregate the representations
of the contextual positions
with higher weights 
for similar representations.

{
\begin{figure}[htb]
\centering
\includegraphics[width=0.8\textwidth]{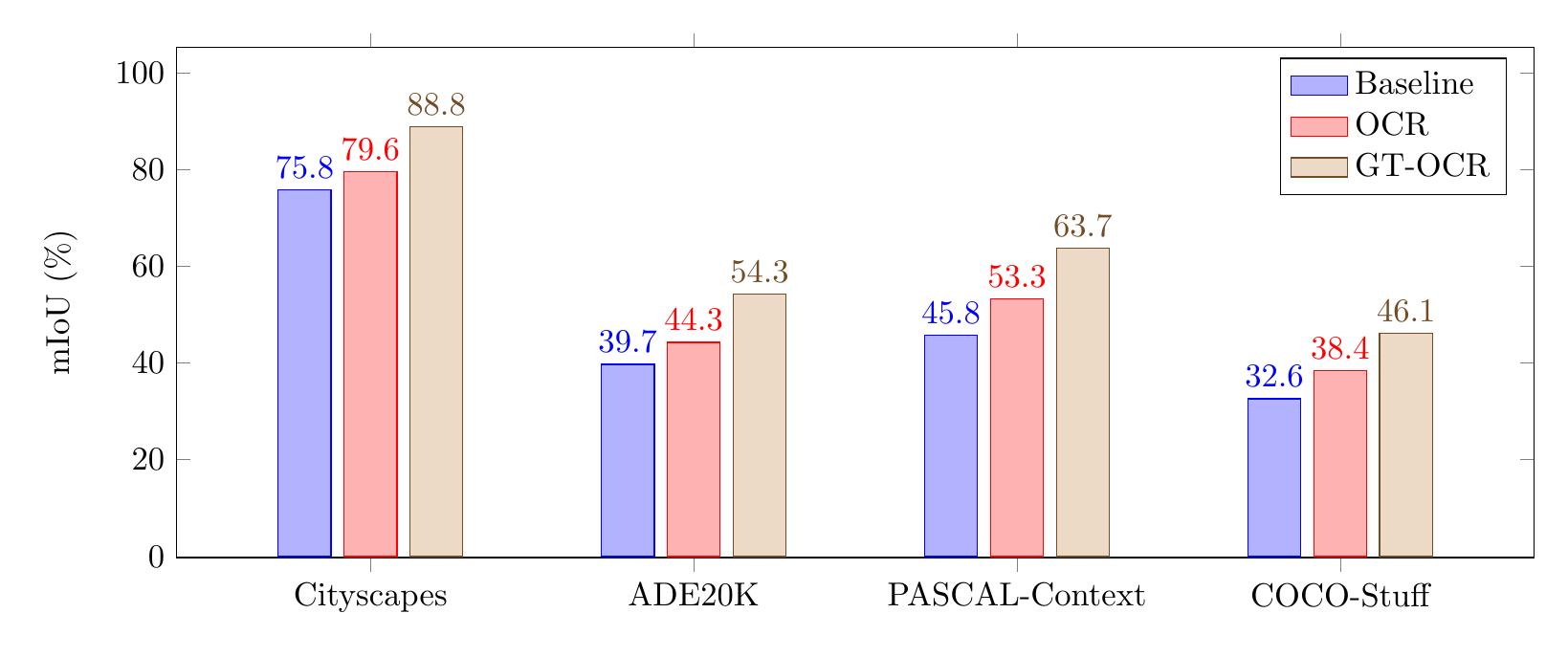}
\caption{\small{
\textbf{
Illustrating the effectiveness of our OCR scheme.
}
GT-OCR estimates the ideal object-contextual representations through
exploiting the ground-truth, which is the upper-bound of our method.
OCR reports the performance of our proposed object-contextual representations.
The three methods,
Baseline, OCR and GT-OCR, 
use the dilated ResNet-$101$ with output stride $8$
as the backbone.
We evaluate their
(single-scale) segmentation results on
Cityscapes \texttt{val}, ADE20K \texttt{val}, PASCAL-Context \texttt{test}
and COCO-Stuff \texttt{test} separately.
}}
\label{fig:ideal_ocr}
\end{figure}
}

We propose to investigate the contextual representation scheme
along the line
of exploring the relation
between a position and its context.
The motivation
is that 
\emph{the class label assigned 
to one pixel is the category of the object
\footnote{
We use ``object" to represent
both ``things" and ``stuff" 
following~\cite{farabet2012learning,shetty2019not}.
}
that the pixel belongs to}.
We aim to augment the representation of one pixel by 
exploiting the representation of 
the object region of the corresponding class.
The empirical study,
shown in Fig.~\ref{fig:ideal_ocr},
verifies
that such a representation augmentation scheme,
when the ground-truth object region is given,
dramatically improves 
the segmentation quality
\footnote{See Section 3.4
for more details.}.

Our approach
consists of three main steps.
First, we divide the contextual pixels
into a set of soft object regions
with each corresponding to a class,
i.e., a coarse soft segmentation computed from a deep network (e.g., ResNet~\cite{he2016deep} or HRNet~\cite{sun2019high}).
Such division is learned under the supervision of the ground-truth segmentation.
Second,
we estimate the representation
for each object region
by aggregating the representations
of the pixels in the corresponding object region.
Last,
we augment
the representation of each pixel
with the object-contextual representation (OCR).
The OCR is the weighted aggregation of all
the object region representations 
with the weights 
calculated according to the relations
between pixels and object regions.

The proposed OCR approach differs
from the conventional multi-scale context schemes.
Our OCR differentiates the same-object-class contextual pixels 
from the different-object-class contextual pixels,
while the multi-scale context schemes,
such as ASPP~\cite{chen2017rethinking} 
and PPM~\cite{zhao2017pyramid},
do not,
and only differentiate the pixels
with different spatial positions.
Fig.~\ref{fig:context_compare} provides
an example to illustrate the differences between 
our OCR context and the multi-scale context.
On the other hand,
our OCR approach is also different 
from the previous relational context schemes~\cite{wang2018non,fu2018dual,yuan2018ocnet,Zhang_2019_ICCV,zhang2019co}.
Our approach structures the
contextual pixels into object regions
and exploits the relations
between pixels and object regions.
In contrast,
the previous relational context schemes
consider the contextual pixels separately
and only exploit the relations
between pixels and contextual pixels~\cite{fu2018dual,yuan2018ocnet,zhang2019co}
or predict the relations
only from pixels without considering the regions~\cite{Zhang_2019_ICCV}.

We evaluate our approach 
on various challenging semantic segmentation benchmarks.
Our approach outperforms the multi-scale context schemes,
e.g., PSPNet, DeepLabv3, 
and 
the recent relational context schemes,
e.g., DANet, 
and the efficiency is also improved.
Our approach achieves competitive performance on 
five benchmarks:
$84.5\%$ on Cityscapes \texttt{test}, 
$45.66\%$ on ADE20K \texttt{val},
$56.65\%$ on LIP \texttt{val},
$56.2\%$ on PASCAL-Context \texttt{test}
and 
$40.5\%$ on COCO-Stuff \texttt{test}.
Besides, we extend our approach to Panoptic-FPN~\cite{kirillov2019panoptic}
and verify the effectiveness of our OCR
on the COCO panoptic segmentation task,
e.g., Panoptic-FPN + OCR achieves $44.2\%$ on COCO \texttt{val}.

\section{Related Work}
\noindent\textbf{Multi-scale context.}
PSPNet~\cite{zhao2017pyramid} 
performs regular convolutions 
on pyramid pooling representations to capture the multi-scale context.
The DeepLab series~\cite{chen2018deeplab,chen2017rethinking}
adopt parallel dilated convolutions
with different dilation rates 
(each rate captures
the context of a different scale).
The recent works~\cite{he2019adaptive,Yang_2018_CVPR,Zhu_2019_ICCV,yuan2018ocnet}
propose various extensions, e.g., DenseASPP~\cite{Yang_2018_CVPR} densifies the dilated rates to cover larger scale ranges.
Some other studies~\cite{chen2018encoder,Lin_2019_CVPR,Fu_2019_ICCV}
construct the encoder-decoder structures to exploit the multi-resolution features as the multi-scale context.

\definecolor{darkpastelpurple}{rgb}{0.59, 0.44, 0.84}
\definecolor{goldenpoppy}{rgb}{0.99, 0.76, 0.0}
\definecolor{apricot}{rgb}{0.98, 0.81, 0.69}
\definecolor{babyblue}{rgb}{0.54, 0.81, 0.94}
\definecolor{carnationpink}{rgb}{1.0, 0.65, 0.79}

\begin{figure}[t]
	\centering
	\begin{subfigure}[b]{0.32\textwidth}
		{\includegraphics[width=\textwidth]{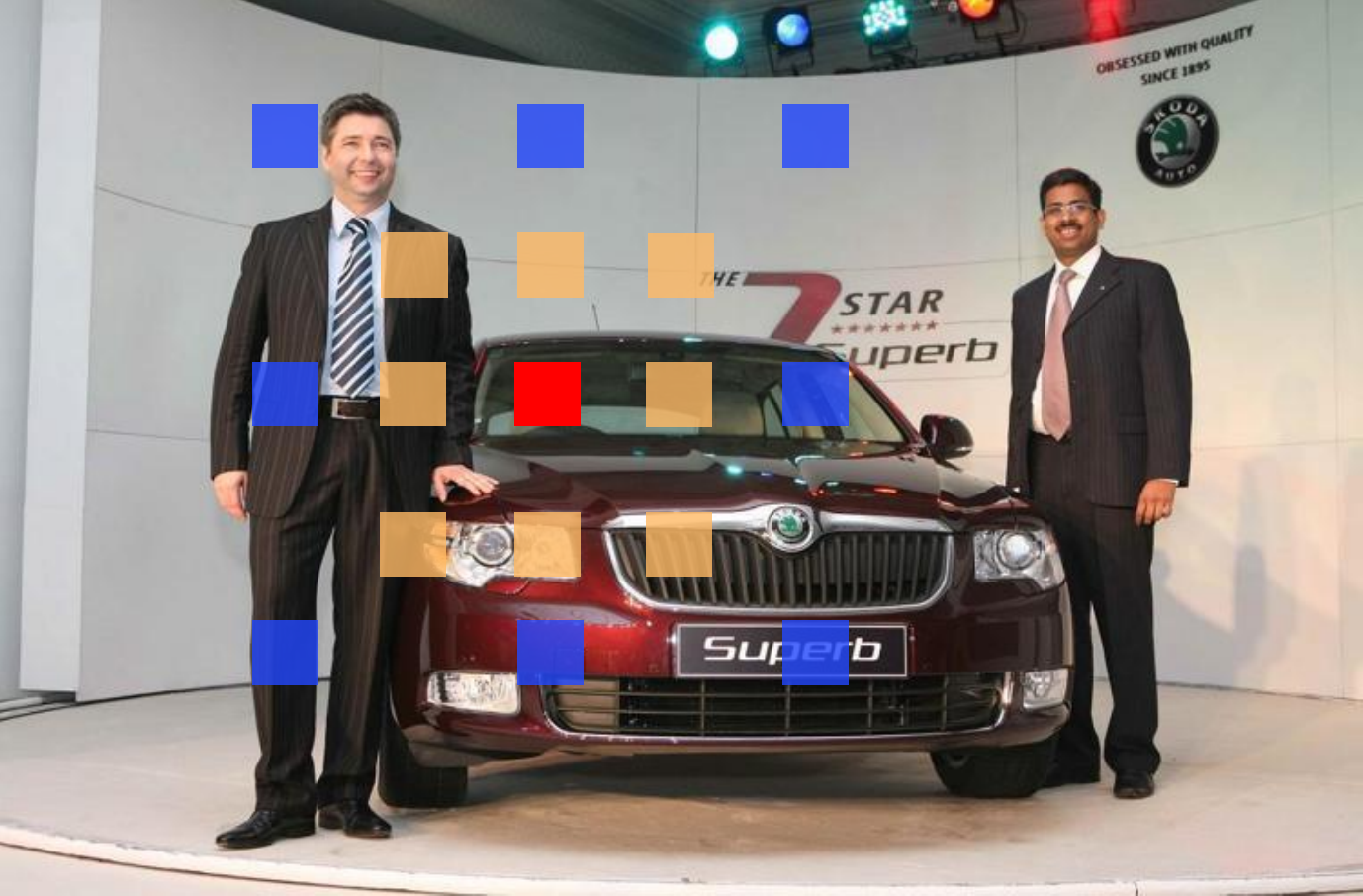}}
		\caption{ASPP}
	\end{subfigure}
	\begin{subfigure}[b]{0.32\textwidth}
		{\includegraphics[width=\textwidth]{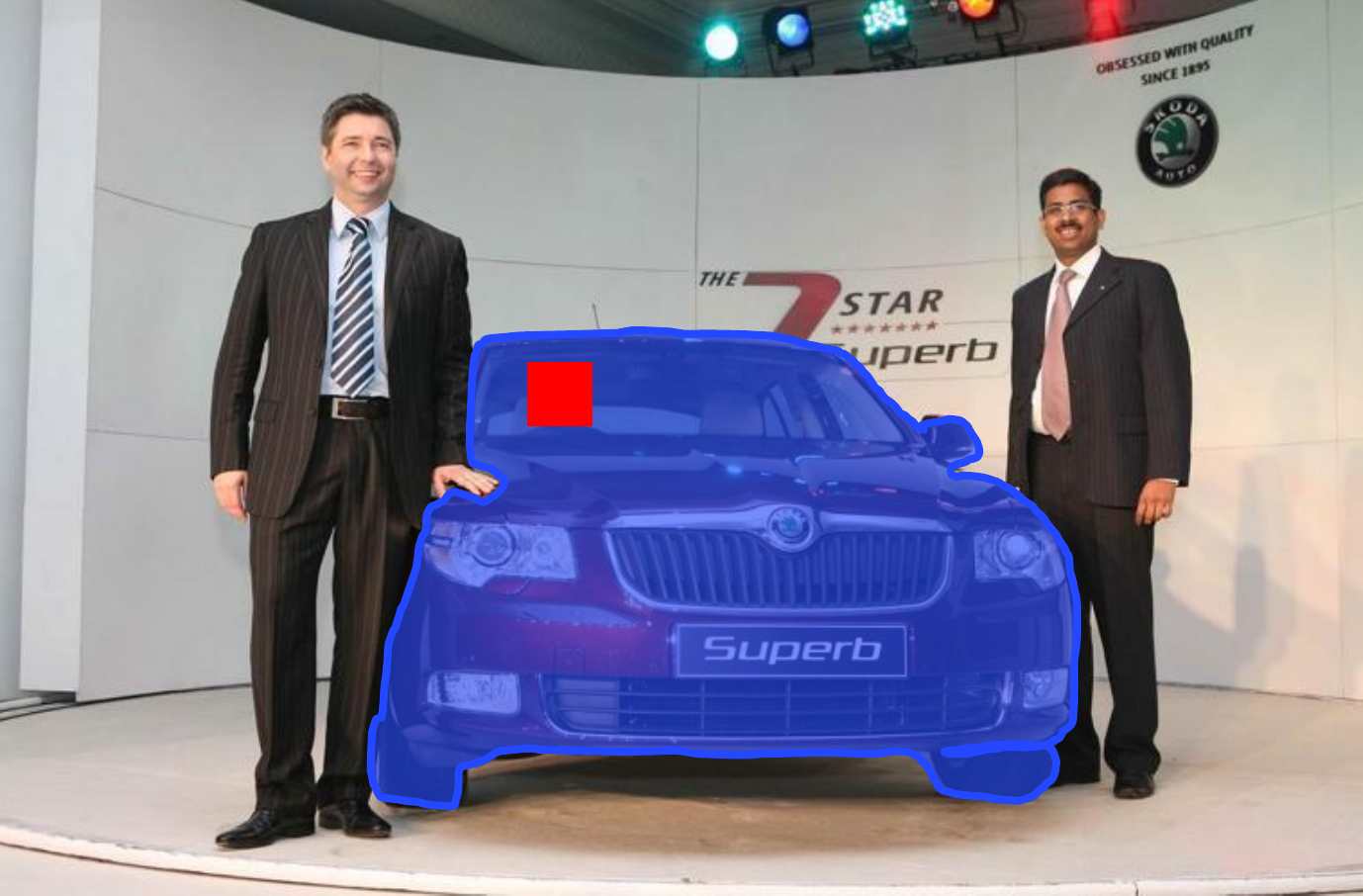}}
		\caption{OCR}
	\end{subfigure}
	\caption{
		\small{
		    \textbf{
			Illustrating the multi-scale context 
			with the ASPP as an example 
			and the OCR context
			for the pixel marked with} {\color{red}{\ding{110}}}.
			(a) ASPP: 
			The context is a set of sparsely sampled pixels
			marked with {\color{apricot}{\ding{110}}},~{\color{blue}{\ding{110}}}.
			The pixels with different colors correspond to 
			different dilation rates.
			Those pixels are distributed 
			in both the object region and the background region.
			(b) Our OCR:
			The context is expected to be
			a set of pixels lying in the object (marked with color \textcolor{blue}{blue}).
			The image is chosen from ADE$20$K.
		}
	}
	\label{fig:context_compare}
\end{figure}

\noindent\textbf{Relational context.}
DANet~\cite{fu2018dual}, CFNet~\cite{zhang2019co} and OCNet~\cite{yuan2018ocnet,yuan2021}
augment the representation
for each pixel
by aggregating the representations
of the contextual pixels,
where the context consists of all the pixels.
Different from the global context~\cite{liu2015parsenet},
these works consider the relation (or similarity) between the pixels,
which is based on the self-attention scheme~\cite{wang2018non,vaswani2017attention},
and perform a weighted aggregation
with the similarities as the weights.

Double Attention 
and its related work~\cite{A2Net,Zhang_2019_ICCV,chen2018graph,NIPS2018_7456,li2018beyond,NIPS2018_7886,li19,huang2019interlaced}
and ACFNet~\cite{Zhang_2019_ICCV}
group the pixels into a set of regions,
and then augment the pixel representations
by aggregating the region representations
with the consideration of their context relations
predicted by using the pixel representation.

Our approach is a relational context approach
and is related to Double Attention and ACFNet.
The differences lie in the region formation and 
the pixel-region relation computation.
Our approach learns the regions with
the supervision of
the ground-truth segmentation.
In contrast,
the regions in previous approaches
except ACFNet
are formed unsupervisedly.
On the other hand,
the relation between a pixel and a region
is computed by considering both the pixel and region representations,
while the relation in previous works is only computed
from the pixel representation.

\noindent\textbf{Coarse-to-fine segmentation.}
Various coarse-to-fine segmentation schemes have been developed~\cite{fieraru2018learning,gidaris2017detect,li2016iterative,tu2010auto,islam2017label,kuo2019shapemask,zhu20183d}
to gradually refine the segmentation maps
from coarse to fine.
For example,~\cite{li2016iterative} 
regards the coarse segmentation map
as an additional representation
and combines it with the original image or other representations
for computing a fine segmentation map.

Our approach in some sense can also be regarded
as a coarse-to-fine scheme.
The difference lies in that we use the coarse segmentation map
for generating a contextual representation
instead of directly used as an extra representation.
We compare our approach with the conventional coarse-to-fine
schemes in the supplementary material.

\noindent\textbf{Region-wise segmentation.} 
There exist many region-wise segmentation methods~\cite{arbelaez2012semantic,caesar2016region,gu2009recognition,gould2009decomposing,wei2017object,neuhold2017mapillary,caesar2016region,uijlings2013selective} that 
organize the pixels into a set of regions (usually super-pixels),
and then classify each region to get the image segmentation result.
Our approach does not classify each region for segmentation
and instead uses the region to learn a better representation for the pixel, which leads to better pixel labeling.

\section{Approach}
Semantic segmentation is a problem of assigning one label $l_i$ 
to each pixel $p_i$ of an image $\mathsf{I}$, 
where $l_i$ is one of $K$ different classes.

\subsection{Background}

\noindent\textbf{Multi-scale context.}
The ASPP~\cite{chen2018deeplab} module captures the multi-scale context information
by performing several parallel dilated convolutions with different dilation rates ~\cite{chen2018deeplab,chen2017rethinking,yu2015multi}:
\begin{align}
\mathbf{y}_i^d
= \sum_{\mathbf{p}_s = \mathbf{p}_i + d \Delta_t} \mathbf{K}^d_t \mathbf{x}_{s}.
\end{align}
Here, $\mathbf{p}_s=\mathbf{p}_i + d \Delta_t$
is the $s$th sampled position
for the dilation convolution with the dilation rate $d$
(e.g., $d=12, 24, 36$ in DeepLabv3~\cite{chen2017rethinking})
at the position $\mathbf{p}_i$.
$t$ is the position index for a convolution,
e.g.,
$\{\Delta_t = (\Delta_w, \Delta_h)|
\Delta_w = -1, 0, 1, 
\Delta_h = -1, 0, 1\}$
for a $3\times 3$ convolution.
$\mathbf{x}_s$ is the representation
at $\mathbf{p}_s$.
$\mathbf{y}_i^d$ is the 
output representation
at $\mathbf{p}_i$
for the $d$th dilated convolution.
$\mathbf{K}^d_t$ is the kernel parameter
at position $t$ for for the $d$th dilated convolution.
The output multi-scale contextual representation
is the concatenation 
of the representations output by the parallel dilated convolutions.

\begin{figure*}[t]
\centering
\resizebox{\linewidth}{!}
{
\includegraphics[scale=0.6]{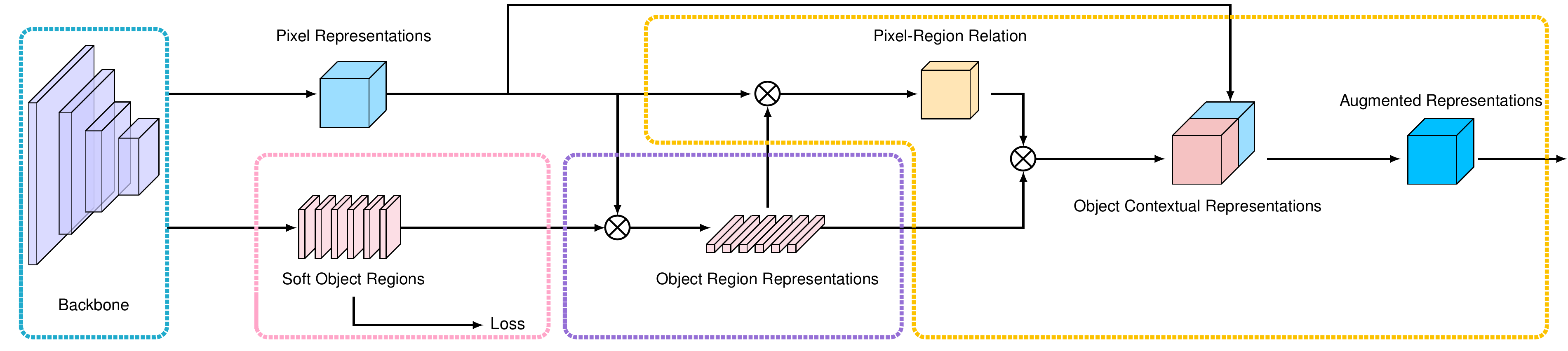}
}
\caption{\small{
\textbf{
Illustrating the pipeline of OCR.
}
(i) form the soft object regions in the \emph{{\color{carnationpink}pink} dashed box}.
(ii) estimate the object region representations in the \emph{{\color{darkpastelpurple}purple} dashed box};
(iii) compute the object contextual representations and the augmented representations 
in the \emph{{\color{goldenpoppy}orange} dashed box}.
See Section 3.2 and 3.3
for more details.
}}
\label{fig:ocr_pipeline}
\end{figure*}

The multi-scale context scheme 
based on dilated convolutions
captures the contexts of multiple scales
without losing the resolution.
The pyramid pooling module in PSPNet~\cite{zhao2017pyramid}
performs regular convolutions
on representations of different scales,
and also captures the contexts of multiple scales
but loses the resolution for large scale contexts.

\noindent\textbf{Relational context.}
The relational context scheme~\cite{fu2018dual,yuan2018ocnet,zhang2019co}
computes the context for each pixel
by considering the relations:
\begin{align}
\mathbf{y}_i = \rho(\sum_{s \in \mathcal{I}}
w_{is} \delta(\mathbf{x}_s)),
\label{eqn:relation_context}
\end{align}
where $\mathcal{I}$
refers to the set of pixels in the image,
$w_{is}$
is the relation 
between $\mathbf{x}_i$
and 
$\mathbf{x}_s$,
and may be predicted
only from $\mathbf{x}_i$
or computed from $\mathbf{x}_i$
and 
$\mathbf{x}_s$.
$\delta(\cdot)$ and $\rho(\cdot)$ are two
different transform functions 
as done in self-attention~\cite{vaswani2017attention}.
The global context scheme~\cite{liu2015parsenet} is 
a special case of relational context with $w_{is}=\frac{1}{|\mathcal{I}|}$.

\subsection{Formulation}
\label{sec:formulation}

The class label $l_i$ for pixel $p_i$ is essentially 
the label of the object that pixel $p_i$ lies in.
Motivated by this,
we present an object-contextual representation approach,
characterizing each pixel by exploiting the corresponding object representation.

The proposed object-contextual representation scheme
(1) structurizes all the pixels in image $\mathsf{I}$
into $K$ soft object regions,
(2) represents each object region 
as $\mathbf{f}_k$
by aggregating the representations
of all the pixels in the $k$th object region,
and (3) augments the representation
for each pixel
by aggregating the $K$ object region representations
with consideration of its relations
with all the object regions:
\begin{align}
\mathbf{y}_i
= \rho(\sum_{k=1}^K w_{ik} \delta(\mathbf{f}_k)),
\label{eqn:ocr}
\end{align}
where $\mathbf{f}_k$ is the representation
of the $k$th object region,
$w_{ik}$ is the relation
between the $i$th pixel
and the $k$th object region.
$\delta(\cdot)$ and $\rho(\cdot)$ 
are transformation functions.

\noindent\textbf{Soft object regions.}
We partition the image $\mathsf{I}$ into $K$ soft 
object regions 
$\{\mathbf{M}_1, \mathbf{M}_2, \dots, \mathbf{M}_K\}$.
Each object region $\mathbf{M}_k$ corresponds to the class $k$,
and is represented by a $2$D map (or coarse segmentation map), where each entry indicates 
the degree that the corresponding pixel belongs 
to the class $k$.

We compute the $K$ object regions
from an intermediate representation output
from a backbone
(e.g., ResNet or HRNet).
During training,
we learn the object region generator
under the supervision 
from the ground-truth segmentation
using the cross-entropy loss.

\noindent\textbf{Object region representations.}
We aggregate
the representations of all the pixels weighted
by their degrees belonging to the $k$th object region,
forming the $k$th object region representation:
\begin{align}
    \mathbf{f}_k = \sum_{i \in \mathcal{I}} {\tilde{m}_{ki}} \mathbf{x}_i. \label{eq:regionrepresentation}
\end{align}
Here, $\mathbf{x}_i$ is the representation of pixel $p_i$.
$\tilde{m}_{ki}$ 
is the normalized degree for pixel $p_i$ belonging to the $k$th object region.
We use spatial softmax to normalize
each object region $\mathbf{M}_k$.

\noindent\textbf{Object contextual representations.}
We compute the relation between each pixel and each object region as below:
\begin{align}
    w_{ik} = \frac{e^{\kappa(\mathbf{x}_i, \mathbf{f}_k)}}
    {\sum_{j=1}^{K} e^{\kappa(\mathbf{x}_i, \mathbf{f}_j)}}. \label{eq:OCRweight}
\end{align}
Here, $\kappa(\mathbf{x}, \mathbf{f}) = \phi(\mathbf{x})^\top\psi(\mathbf{f})$
is the unnormalized relation function,
$\phi(\cdot)$ and $\psi(\cdot)$
are two transformation functions implemented by
$1\times 1~\operatorname{conv} \rightarrow \operatorname{BN} \rightarrow \operatorname{ReLU}$.
This is inspired by self-attention~\cite{vaswani2017attention}
for a better relation estimation.

The object contextual representation $\mathbf{y}_i$ for pixel $p_i$ 
is computed 
according to Equation~\ref{eqn:ocr}.
In this equation, 
$\delta(\cdot)$ and $\rho(\cdot)$
are both transformation functions implemented by
$1\times 1~\operatorname{conv} \rightarrow \operatorname{BN} \rightarrow \operatorname{ReLU}$,
and this follows non-local networks~\cite{wang2018non}.

\noindent\textbf{Augmented representations.}
The final representation for pixel $p_i$ 
is updated as the aggregation of two parts,
(1) the original representation $\mathbf{x}_i$,
and (2) the object contextual representation $\mathbf{y}_i$:
\begin{align}
    \mathbf{z}_i = g([\mathbf{x}_i^\top~\mathbf{y}_i^\top]^\top). \label{eq:eq4}
\end{align}
where $g(\cdot)$ is a transform function used to fuse the original representation
and the object contextual representation, implemented by $1\times 1~\operatorname{conv} \rightarrow \operatorname{BN} \rightarrow \operatorname{ReLU}$.
The whole pipeline of our approach 
is illustrated in Fig.~\ref{fig:ocr_pipeline}.

\noindent\emph{Comments:}
Some recent studies,
e.g., Double Attention~\cite{A2Net} 
and ACFNet~\cite{Zhang_2019_ICCV},
can be formulated similarly to Equation~\ref{eqn:ocr},
but differ from our approach
in some aspects.
For example, 
the region formed in Double Attention 
do not correspond to an object class,
and the relation in ACFNet~\cite{Zhang_2019_ICCV}
is computed only from the pixel representation
w/o using the object region representation.

{\color{darkgray}
\begin{figure*}[t]
\centering
    \includegraphics{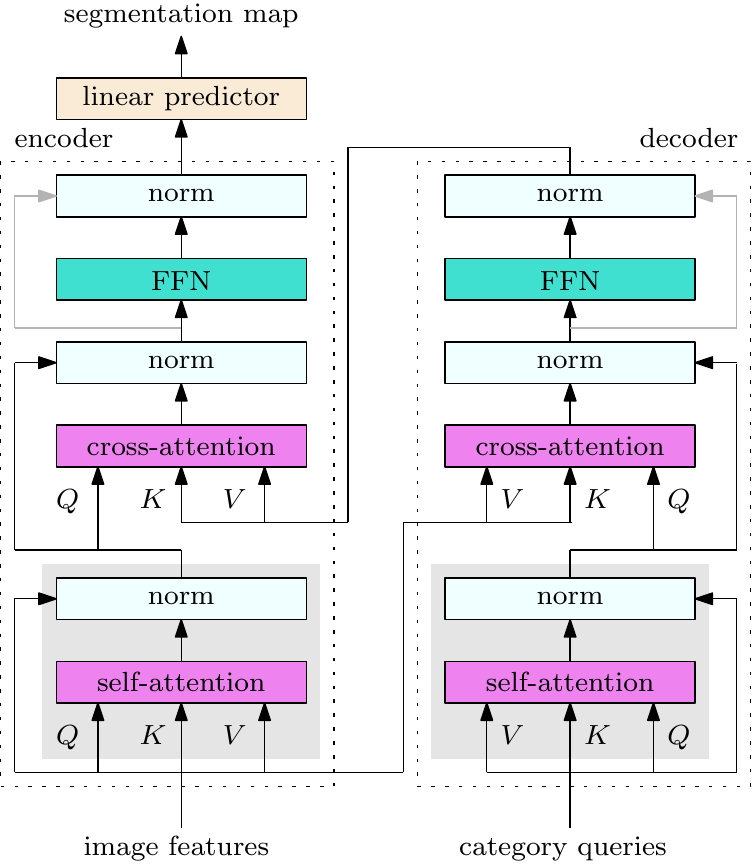} 
    \caption{\small
    \textbf{Segmentation transformer}. Rephrasing the OCR pipeline showing
    in Figure~\ref{fig:ocr_pipeline} 
    using the Transformer encoder-decoder architecture.
    The encoder self-attention unit in the gray box could be a local version and 
    is optional
    and is useful for boosting the image features.
    The decoder self-attention unit serves as the role
    of interacting the category queries
    and can be discarded for a single decoder layer
    or moved after the decoder cross-attention unit.}
    \label{fig:segmentationtransformer}
\end{figure*}

\subsection{Segmentation Transformer: Rephrasing the OCR Method}
\label{sec:segmentationtransformer}
We rephrase the OCR pipeline using the Transformer~\cite{VaswaniSPUJGKP17} language
and illustrate the Transformer encoder-decoder architecture
in Figure~\ref{fig:segmentationtransformer}. 
The aforementioned OCR pipeline
consists of three steps:
soft object region extraction,
object region representation computation,
and object-contextual representation computation for each position,
and mainly explores
the decoder and encoder cross-attention modules.

\vspace{0.1cm}
\noindent\textbf{Attention.}
The attention~\cite{VaswaniSPUJGKP17}
is computed using
the scaled dot-product. 
The inputs contain:
a set of $N_q$ queries $\mathbf{Q} \in \mathbb{R}^{d \times N_q}$,
a set of $N_{kv}$ keys $\mathbf{K} \in \mathbb{R}^{d \times N_{kv}}$,
and a set of $N_{kv}$ values $\mathbf{V} \in \mathbb{R}^{d \times N_{kv}}$.
The attention weight $a_{ij}$
is computed as 
the softmax normalization of
the dot-product between the query $\mathbf{q}_{i}$ 
and the key $\mathbf{k}_{j}$:
\begin{align}
    a_{ij} = \frac{e^{\frac{1}{\sqrt{d}}\mathbf{q}_{i}^\top \mathbf{k}_{j}}}{Z_i}~~
    \text{where~~} Z_i = \sum\nolimits_{j=1}^{N_{kv}} e^{\frac{1}{\sqrt{d}}\mathbf{q}_{i}^\top \mathbf{k}_{j}}.
    \label{eqn:attentionweight}
\end{align}
The attention output 
for each query $\mathbf{q}_{i}$ is the aggregation 
of values weighted by attention weights:
\begin{align}
    \operatorname{Attn}(\mathbf{q}_{i}, 
    \mathbf{K},
    \mathbf{V})
    = \sum\nolimits_{j=1}^{N_{kv}} \alpha_{ij}\mathbf{v}_{j}.
    \label{eqn:attentionfeature}
\end{align}

\vspace{0.1cm}
\noindent\textbf{Decoder cross-attention.}
The decoder cross-attention module has two roles:
soft object region extraction and
object region representation computation.

The keys and values are image features
($\mathbf{x}_i$ in Equation~\ref{eq:regionrepresentation}).
The queries are $K$ category queries
($\mathbf{q}_1, \mathbf{q}_2, \dots, \mathbf{q}_K$),
each of which corresponds to a category.
The $K$ category queries essentially are used
to generate the soft object regions,
$\mathbf{M}_1, \mathbf{M}_2, \dots, \mathbf{M}_K$, 
which are later spatially softmax-normalized as the weights 
$\tilde{m}$ in Equation~\ref{eq:regionrepresentation}. 
Computing $\tilde{m}$
is the same as the manner of computing 
attention the weight $\alpha_{ij}$
in Equation~\ref{eqn:attentionweight}.
The object region representation computation manner in Equation~\ref{eq:regionrepresentation}
is the same as Equation~\ref{eqn:attentionfeature}.

\vspace{0.1cm}
\noindent\textbf{Encoder cross-attention.}
The encoder cross-attention module (with the subsequent FFN) serves as the role 
of aggregating the object region representations
as shown in Equation~\ref{eqn:ocr}.
The queries are image features at each position,
and the keys and values are the decoder outputs. 
Equation~\ref{eq:OCRweight}
computes the weights
in a way the same as
the attention computation manner Equation~\ref{eqn:attentionweight},
and the contextual aggregation Equation~\ref{eqn:ocr}
is the same as Equation~\ref{eqn:attentionfeature}
and $\rho(\cdot)$ corresponds to the $\operatorname{FFN}$ operator.

\vspace{0.1cm}
\noindent\textbf{Connection to class embedding and class attention~\cite{DosovitskiyBKWZUDMHGUH20,TouvronCSSJ21}.}
The category queries
are close to the class embedding 
in Vision Transformer (ViT)~\cite{DosovitskiyBKWZUDMHGUH20} and in Class-Attention in Image Transformers (CaiT)~\cite{TouvronCSSJ21}.
We have an embedding for each class
other than an integrated embedding for all the classes.
The decoder cross attention in segmentation transformer is similar to class attention 
in CaiT~\cite{TouvronCSSJ21}.

The encoder and decoder architecture
is close to self-attention in ViT
over both the class embedding and image features.
If the two cross-attentions and the two self-attentions are conducted simultaneously
(depicted in Figure~\ref{fig:segmentationtransformerv2}),
it is equivalent to a single self-attention. 
It is interesting 
to learn the attention parameters for category queries
at the ImageNet pre-training stage.

\begin{figure}
    \centering
    \includegraphics{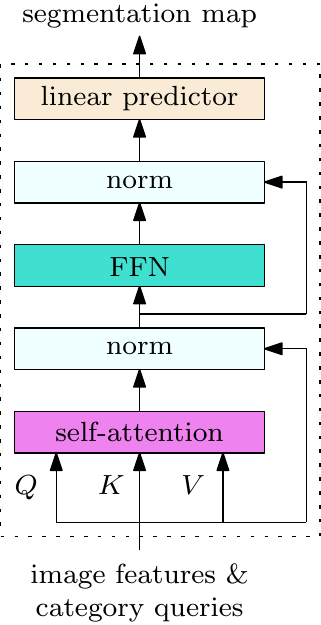}
    \caption{\small\textbf{An alternative of
    segmentation transformer} (shown in Figure~\ref{fig:segmentationtransformer}).
    The module in the dashed box is repeated several times.}
    \label{fig:segmentationtransformerv2}
\end{figure}

\vspace{0.1cm}
\noindent\textbf{Connection to OCNet and interlaced self-attention~\cite{yuan2021}.}
The OCNet~\cite{yuan2021} exploits the self-attention 
(i.e., only the encoder self-attention unit is included in Figure~\ref{fig:segmentationtransformer},
and the encoder cross-attention unit and the decoder are not included).
The self-attention unit is accelerated by
an interlaced self-attention unit,
consisting of local self-attention and global self-attention
that can be simplified as self-attention over the pooled features over
the local windows\footnote{The local and/or global self-attention units in interlaced self-attention~\cite{yuan2021}
could be applied to Vision Transformer~\cite{2020vit}
for acceleration.}.
As an alternative scheme,
the category queries in Figure~\ref{fig:segmentationtransformer}
could be replaced 
by regularly-sampled or adaptively-pooled image features
other than learned as model parameters.

}

\subsection{Architecture}

\noindent\textbf{Backbone.}
We use the dilated ResNet-$101$~\cite{he2016deep} (with output stride $8$) or HRNet-W$48$~\cite{sun2019high} (with output stride $4$) as the backbone.
For dilated ResNet-$101$, 
there are two representations input to the OCR module.
The first representation from Stage $3$ 
is for predicting coarse segmentation (object regions).
The other representation from Stage $4$
goes through a $3\times3$ convolution ($512$ output channels),
and then is fed into the OCR module.
For HRNet-W$48$, we only use the final representation as the input to 
the OCR module.

\noindent\textbf{OCR module.}
We implement the above formulation of our approach as the OCR module,
as illustrated in Fig.~\ref{fig:ocr_pipeline}.
We use a linear function (a $1\times 1$ convolution)
to predict the coarse segmentation (soft object region) supervised with a
pixel-wise cross-entropy loss. 
All the transform functions,
$\psi(\cdot),~\phi(\cdot),~\delta(\cdot),~\rho(\cdot),~\text{and}~g(\cdot)$,
are implemented as 
$1\times 1~\operatorname{conv} \rightarrow \operatorname{BN} \rightarrow \operatorname{ReLU}$,
and the first three output $256$ channels
and the last two output $512$ channels.
We predict the final segmentation from the final representation
using a linear function and we also apply a pixel-wise cross-entropy loss
on the final segmentation prediction.

\subsection{Empirical Analysis}

We conduct the empirical analysis experiments using
the dilated ResNet-$101$ as
the backbone
on Cityscapes \texttt{val}.

\noindent\textbf{Object region supervision.}
We study the influence of the object region supervision.
We modify our approach through removing the supervision (i.e., loss)
on the soft object regions (within the {\color{carnationpink}pink} dashed box in Fig.~\ref{fig:ocr_pipeline}),
and adding 
another auxiliary loss in the stage-$3$ of ResNet-$101$.
We keep all the other settings the same and report the results in
the left-most $2$ columns of Table~\ref{table:pixel_region_relation}.
We can see that the supervision for forming the object regions
is crucial for the performance.

\noindent\textbf{Pixel-region relations.}
We compare our approach with other two mechanisms
that do not use the region representation
for estimating the pixel-region relations:
(i) Double-Attention~\cite{A2Net}
uses the pixel representation to
predict the relation;
(ii) ACFNet~\cite{Zhang_2019_ICCV}
directly uses one intermediate segmentation map
to indicate the relations.
We use DA scheme and ACF scheme to represent the
above two mechanisms.
We implement both methods by ourselves
and only use the dilated ResNet-$101$ as
the backbone
without using multi-scale contexts
(the results of ACFNet
is improved by using ASPP~\cite{Zhang_2019_ICCV})

\renewcommand{\arraystretch}{1}
\begin{table}
\centering
\small
\caption{\small{
\textbf{Influence of object region supervision and pixel-region relation estimation scheme.}
We can find both the object region supervision
and our pixel-region relation scheme are important
for the performance.
}}
\resizebox{0.9\linewidth}{!}
{
\begin{tabular}{>{\centering\arraybackslash}p{2.5cm}|>{\centering\arraybackslash}p{2.5cm}|>{\centering\arraybackslash}p{2.5cm}|>{\centering\arraybackslash}p{2.5cm}|>{\centering\arraybackslash}p{2cm}}
    \shline
    \multicolumn{2}{c|}{Object region supervision} & \multicolumn{3}{c}{Pixel-region relations} \\
    \hline
    w/o supervision & w/ supervision & DA scheme  & ACF scheme & Ours \\ 
    \hline
    $77.31\%$ & $\bf{79.58}\%$ & $79.01\%$ & $78.02\%$  & $\bf{79.58}\%$ \\
    \shline
\end{tabular}
}
\label{table:pixel_region_relation}
\end{table}

The comparison in Table~\ref{table:pixel_region_relation}
shows that our approach gets superior performance.
The reason is that 
we exploit the pixel representation
as well as the region representation 
for computing the relations.
The region representation
is able to 
characterize the object 
in the specific image,
and thus the relation is more accurate
for the specific image
than that 
only using the pixel representation.

\noindent\textbf{Ground-truth OCR.}
We study the segmentation performance
using the ground-truth segmentation
to form the object regions and
the pixel-region relations, called GT-OCR,
to justify our motivation.
(i) Object region formation using the ground-truth: 
set the confidence of pixel $i$ belonging to $k$th object region 
$m_{ki}=1$ if the ground-truth label $l_i\equiv k$ and $m_{ki}=0$ otherwise.
(ii) Pixel-region relation computation using the ground-truth:
set the pixel-region relation $w_{ik}=1$ if the 
ground-truth label $l_i\equiv k$
and $w_{ik}=0$ otherwise.
We have illustrated the detailed results of 
GT-OCR on four different benchmarks in Fig.~\ref{fig:ideal_ocr}.

\section{Experiments: Semantic Segmentation}

\subsection{Datasets}
\noindent\textbf{Cityscapes}. 
The Cityscapes dataset~\cite{cordts2016cityscapes} is tasked for urban scene understanding. 
There are totally
$30$ classes 
and only $19$ classes are used for parsing evaluation.
The dataset contains $5$K high quality pixel-level finely annotated images and $20$K coarsely annotated images. The finely annotated $5$K images are divided into $2,975/500/1,525$ images for training, validation and testing.

\noindent\textbf{ADE20K}.
The ADE$20$K dataset~\cite{zhou2017scene} is used in ImageNet scene parsing challenge 2016.
There are $150$ classes and diverse scenes with $1,038$ image-level labels.
The dataset is divided into $20$K/$2$K/$3$K images for training, validation and testing. 

\noindent\textbf{LIP}.
The LIP dataset~\cite{Gong_2017_CVPR} is used in the LIP challenge 2016 for single human parsing task. 
There are about $50$K images with $20$ classes ($19$ semantic human part classes and $1$ background class).
The training, validation, and test sets consist of 
$30$K, $10$K, $10$K images respectively.

\noindent\textbf{PASCAL-Context}.
The PASCAL-Context dataset~\cite{mottaghi2014role} is a challenging
scene parsing dataset that contains $59$ semantic
classes and $1$ background class. 
The training set and test set consist of
$4,998$ and $5,105$ images respectively.

\noindent\textbf{COCO-Stuff}.
The COCO-Stuff dataset~\cite{caesar2018coco} is a challenging
scene parsing dataset that contains $171$ semantic
classes. 
The training set and test set consist of
$9$K and $1$K images respectively.

\subsection{Implementation Details}

\noindent\textbf{Training setting.}
We initialize the backbones
using the model pre-trained on ImageNet
and the OCR module randomly.
We perform the polynomial learning rate policy with factor $(1-(\frac{iter}{iter_{max}})^{0.9})$, the weight on the final loss as $1$, the weight on the loss used to supervise the object region estimation (or auxiliary loss) as $0.4$.
We use \bnInplaceSync~\cite{Bulo_2018_CVPR} to synchronize the mean and standard-deviation of BN across multiple GPUs.
For the data augmentation, we perform random flipping horizontally, random scaling in the range of $[0.5, 2]$ and random brightness jittering within the range of $[-10, 10]$.
We perform the same training settings for the reproduced approaches, e.g., PPM, ASPP, to ensure the fairness.
We follow the previous works~\cite{chen2017rethinking,Zhang_2018_CVPR,zhao2017pyramid} 
for setting up the training
for the benchmark datasets.

\noindent$\Box$ \emph{Cityscapes}: 
We set the initial learning rate as $0.01$, weight decay as $0.0005$, crop size as $769\times 769$ and batch size as $8$ by default.
For the experiments evaluated on \texttt{val}/\texttt{test} set,
we set training iterations as $40$K/$100$K on \texttt{train}/\texttt{train}+\texttt{val}
set separately.
For the experiments augmented with extra data:
(i) w/ \texttt{coarse},
we first train our model on \texttt{train} + \texttt{val} for $100$K iterations
with initial learning rate as $0.01$,
then we fine-tune the model on \texttt{coarse} set
for $50$K iterations and continue fine-tune our model on
\texttt{train}+\texttt{val} for $20$K iterations with
the same initial learning rate $0.001$.
(ii) w/ \texttt{coarse} + Mapillary~\cite{neuhold2017mapillary},
we first pre-train our model on the Mapillary \texttt{train} set
for $500$K iterations with batch size $16$ and initial learning rate $0.01$ (achieves $50.8\%$ on Mapillary \texttt{val}),
then we fine-tune the model on Cityscapes following the order of \texttt{train} + \texttt{val} ($100$K iterations) $\to$ \texttt{coarse} ($50$K iterations) $\to$ \texttt{train} + \texttt{val} ($20$K iterations),
we set the initial learning rate as $0.001$ and the batch size as $8$
during the above three fine-tuning stages on Cityscapes.

\noindent$\Box$ \emph{ADE20K}: 
We set the initial learning rate as $0.02$, weight decay as $0.0001$, crop size as $520\times 520$, batch size as $16$ and and training iterations as $150$K if not specified.

\noindent$\Box$ \emph{LIP}: 
We set the initial learning rate as $0.007$, weight decay as $0.0005$, crop size as $473\times 473$, batch size as $32$ and training iterations as $100$K if not specified.

\noindent$\Box$ \emph{PASCAL-Context}: We set the initial learning rate as $0.001$, weight decay as $0.0001$, crop size as $520\times 520$, batch size as $16$ and training iterations as $30$K if not specified.

\noindent$\Box$ \emph{COCO-Stuff}: We set the initial learning rate as $0.001$, weight decay as $0.0001$, crop size as $520\times 520$, batch size as $16$ and training iterations as $60$K if not specified.

\subsection{Comparison with Existing Context Schemes}
We conduct the experiments using the dilated ResNet-$101$ as the backbone and use the same training/testing settings to ensure the fairness.

\begin{table}[t]
\begin{minipage}[t]{1\linewidth}
\centering
\footnotesize
\caption{\small{
\textbf{Comparison with multi-scale context scheme.}
We use {$\bigstar$} to mark the result w/o using Cityscapes \texttt{val} for training.
We can find OCR consistently outperforms both PPM and ASPP
across different benchmarks under the fair comparisons.
}}
\resizebox{1\linewidth}{!}
{
\begin{tabular}{>{\arraybackslash}p{5cm}|c|c|>{\centering\arraybackslash}p{2cm}|>{\centering\arraybackslash}p{2cm}}
\shline
Method           & \pbox{40cm}{ Cityscapes (w/o coarse) } & \pbox
{40cm}{ Cityscapes (w/ coarse) }&  \pbox{40cm}{ADE${20}$K}  & \pbox{40cm}{LIP}   \\
\shline
PPM~\cite{zhao2017pyramid}       &   $78.4\%^\bigstar$    &   $81.2\%$  &  $43.29\%$  &  $-$ \\ 
ASPP~\cite{chen2017rethinking}   &   $-$         &   $81.3\%$  &  $-$        &  $-$ \\ 
PPM (Our impl.)                  &   $80.3\%$    &   $81.6\%$  &  $44.50\%$  &  $54.76\%$ \\ 
ASPP (Our impl.)                 &   $81.0\%$    &   $81.7\%$  &  $44.60\%$  &  $55.01\%$ \\ 
\rowcolor{Gray}
OCR                              & $\bf{81.8\%}$ & $\bf{82.4\%}$ &  $\bf{45.28\%}$ &  $\bf{55.60\%}$ \\ 
\shline
\end{tabular}
}
\label{table:ocr_vs_ms}
\end{minipage}
\begin{minipage}[t]{1\linewidth}
\centering
\footnotesize
\caption{\small{
\textbf{Comparison with relational context scheme.}
Our method consistently performs better across different benchmarks.
Notably, Double Attention is sensitive to the region number choice
and we have fine-tuned this hyper-parameter as $64$ to 
report its best performance.
}}
\resizebox{1\linewidth}{!}
{
\begin{tabular}{>{\arraybackslash}p{5cm}|c|c|>{\centering\arraybackslash}p{2cm}|>{\centering\arraybackslash}p{2cm}}
\shline
Method           & \pbox{20cm}{ Cityscapes (w/o coarse) } & \pbox{20cm}{ Cityscapes (w/ coarse) }&  ADE${20}$K  & LIP   \\

\shline
CC-Attention~\cite{Huang_2019_ICCV} &   $81.4\%$  &   -  &  $45.22\%$  &  - \\ 
DANet~\cite{fu2018dual}      &   $81.5\%$  &   -  &  -  &  - \\ 
Self Attention (Our impl.)       &   $81.1\%$    &   $82.0\%$  &  $44.75\%$  &  $55.15\%$ \\ 
Double Attention (Our impl.)     &   $81.2\%$    &   $82.0\%$  &  $44.81\%$  &  $55.12\%$ \\ 
\rowcolor{Gray}
OCR                              & $\bf{81.8\%}$ & $\bf{82.4\%}$ &  $\bf{45.28\%}$ &  $\bf{55.60\%}$ \\ 
\shline
\end{tabular}
}
\label{table:ocr_vs_relation}
\end{minipage}
\begin{minipage}[t]{1\linewidth}
\centering
\footnotesize
\caption{\small{
\textbf{Complexity comparison.} 
We use input feature map of size  [$1\times2048\times128\times128$] 
to evaluate their complexity during inference. 
The numbers are obtained on a single P$40$ GPU with CUDA $10.0$.
All the numbers are the smaller the better.
Our OCR requires the least GPU memory
and the least runtime.}}
\resizebox{0.85\linewidth}{!}
{
\begin{tabular}{>{\arraybackslash}p{5cm}|>{\centering\arraybackslash}p{2cm}|>{\centering\arraybackslash}p{2cm}|>{\centering\arraybackslash}p{2cm}|>{\centering\arraybackslash}p{2cm}}
\shline
Method & Parameters$\blacktriangle$  &  Memory$\blacktriangle$  & FLOPs $\blacktriangle$  & Time$\blacktriangle$ \\
\shline
PPM (Our impl.)   & $23.1$M  & $792$M & $619$G & $99$ms  \\
ASPP (Our impl.)   & $15.5$M  & $284$M  & $492$G & $97$ms  \\
DANet (Our impl.)   & $10.6$M  & $2339$M & $1110$G & $121$ms  \\
CC-Attention (Our impl.)   & $10.6$M  & $427$M & $804$G & $131$ms  \\
Self-Attention (Our impl.)   & ${10.5}$M  & $2168$M & $619$G & $96$ms  \\
Double Attention (Our impl.)     & $\bf{10.2}$M  & $209$M & $\bf{338}$G & $46$ms  \\
\rowcolor{Gray}
OCR   & $10.5$M  & $\bf{202}$M  & $340$G & $\bf{45}$ms  \\
\shline
\end{tabular}
}
\label{table:gpu_compare}
\end{minipage}
\end{table}

\noindent\textbf{Multi-scale contexts.}
We compare our OCR with the multi-scale context schemes including PPM~\cite{zhao2017pyramid} and ASPP~\cite{chen2017rethinking} on three benchmarks including Cityscapes \texttt{test}, ADE$20$K \texttt{val} and LIP \texttt{val} in Table~\ref{table:ocr_vs_ms}.
Our reproduced PPM/ASPP outperforms 
the originally reported numbers in~\cite{zhao2017pyramid,chen2017rethinking}.
From Table~\ref{table:ocr_vs_ms}, 
it can be seen that our OCR outperforms
both multi-scale context schemes by a large margin.
For example, the absolute gains of OCR over PPM (ASPP) for the four comparisons
are 1.5\% (0.8\%),
0.8\% (0.7\%),
0.78\% (0.68\%),
0.84\% (0.5\%).
To the best of our knowledge,
these improvements are already significant
considering that
the baselines (with dilated ResNet-$101$) are already strong
and the complexity of our OCR is much smaller.

\noindent\textbf{Relational contexts.}
We compare our OCR with various relational context schemes including
Self-Attention~\cite{vaswani2017attention,wang2018non}, Criss-Cross attention~\cite{Huang_2019_ICCV} (CC-Attention), DANet~\cite{fu2018dual} and Double Attention~\cite{A2Net} on the same 
three benchmarks including Cityscapes \texttt{test}, ADE$20$K \texttt{val} and LIP \texttt{val}.
For the reproduced Double Attention, we fine-tune the number of the regions (as it is very sensitive to the hyper-parameter choice) and we choose $64$ with the best performance.
More detailed analysis and comparisons are illustrated in the supplementary material.
According to the results in Table~\ref{table:ocr_vs_relation},
it can be seen that our OCR outperforms these relational context schemes
under the fair comparisons.
Notably, the complexity of our OCR is much smaller than most of the other methods.


\noindent\textbf{Complexity.}
We compare the efficiency of our OCR with the efficiencies of the multi-scale context schemes and the relational context schemes.
We measure the increased parameters, GPU memory, computation complexity (measured by the number of FLOPs) and inference time
that are introduced by the context modules,
and do not count the complexity from the backbones.
The comparison in Table~\ref{table:gpu_compare}
shows the superiority of the proposed OCR scheme.

\noindent$\Box$ \emph{Parameters}:
Most relational context schemes require less parameters compared with the multi-scale context schemes.
For example, our OCR only requires 1/2 and 2/3 of the parameters of PPM and ASPP separately.

\noindent$\Box$ \emph{Memory}:
Both our OCR and Double Attention require much less GPU memory compared with the other approaches (e.g., DANet, PPM).
For example, 
our GPU memory consumption is 1/4, 1/10, 1/2, 1/10 of the memory consumption of 
PPM, DANet, CC-Attention
and Self-Attention separately.

\noindent$\Box$ \emph{FLOPs}:
Our OCR only requires 1/2, 7/10, 3/10, 2/5 and 1/2 of the FLOPs based on
PPM, ASPP, DANet, CC-Attention and Self-Attention separately.

\noindent$\Box$ \emph{Running time}:
The runtime of OCR is very small: only 1/2, 1/2, 1/3, 1/3 and 1/2 of
the runtime with PPM, ASPP, DANet, CC-Attention and Self-Attention separately.

In general,
\emph{
our OCR is a much better choice
if we consider the balance between
performance, memory complexity, GFLOPs and running time.
}

\subsection{Comparison with the State-of-the-Art}
Considering that different approaches perform improvements on different baselines to achieve the best performance,
we categorize the existing works to two groups according to the baselines that they apply: (i) \emph{simple baseline:} dilated ResNet-$101$ with stride $8$;
(ii) \emph{advanced baseline:} PSPNet, DeepLabv3, multi-grid (MG), encoder-decoder structures that achieve higher resolution outputs with stride $4$ or stronger backbones such as WideResNet-$38$, Xception-$71$ and HRNet.

For fair comparison with the two groups fairly,
we perform our OCR on a simple baseline (dilated ResNet-$101$ with stride $8$)
and an advanced baseline (HRNet-W$48$ with stride $4$).
Notably, our improvement with HRNet-W$48$ (over ResNet-$101$) is comparable with 
the gain of the other work based on advanced baseline methods.
For example, DGCNet~\cite{zhang2019dual} gains $0.7\%$ with Multi-grid
while OCR gains $0.6\%$ with stronger backbone on Cityscapes \texttt{test}.
We summarize all the results in Table~\ref{table:ocnet_sota_exp_all}
and illustrate the comparison details on each benchmark separately as follows.

\noindent\textbf{Cityscapes.}
Compared with the methods based on 
the simple baseline on Cityscape \texttt{test}
w/o using the coarse data,
our approach achieves the best performance $81.8\%$, which is already 
comparable with some methods based on the advanced baselines, e.g, DANet, ACFNet.
Our approach achieves better performance $82.4\%$ through exploiting 
the coarsely annotated images for training.

\renewcommand{\arraystretch}{1.2}
\begin{table*}[ht!]
\definecolor{darkcoral}{rgb}{0.01, 0.75, 0.24}
\centering
\caption{\small{
\textbf{Comparison with the state-of-the-art.}
We use M to represent multi-scale context and R to represent relational context.
\textcolor{red}{Red}, \textcolor{darkcoral}{Green}, \textcolor{blue}{Blue} represent the top-$3$ results.
We use {$\flat$}, {$\dag$} and {$\ddag$} to mark 
the result w/o using Cityscapes \texttt{val},
the method using Mapillary dataset and 
the method using the Cityscapes video dataset separately
}
}

\resizebox{1\linewidth}{!}{
    \begin{tabular}{r|c|c|c|>{\centering\arraybackslash}p{1.6cm}|>{\centering\arraybackslash}p{1.6cm}|>{\centering\arraybackslash}p{1.6cm}|>{\centering\arraybackslash}p{1.5cm}|>{\centering\arraybackslash}p{1.5cm}|>{\centering\arraybackslash}p{1.6cm}}
    \shline
     Method & Baseline & \pbox{30cm}{Stride} & \pbox{30cm}{Context \\ schemes} & \pbox{30cm}{ Cityscapes\\(w/o coarse)}  &  \pbox{30cm}{ Cityscapes\\(w/ coarse)} & \pbox{30cm}{ADE$20$K} & \pbox{30cm}{LIP}  & \pbox{30cm}{PASCAL \\ Context} & \pbox{30cm}{COCO-Stuff}\\
    \shline
    \multicolumn{10}{c}{Simple baselines} \\
    \hline
    PSPNet~\cite{zhao2017pyramid}  &   ResNet-$101$ & $8\times$  & M & $78.4^{\flat}$ & $81.2$ & $43.29$ & - & \textbf{$47.8$} & - \\ 
    DeepLabv3~\cite{chen2017rethinking} &  ResNet-$101$ & $8\times$ & M & - &  \textcolor{blue}{$81.3$}  & - & - & - & - \\ 
    PSANet~\cite{psanet}           &   ResNet-$101$ & $8\times$ & R & $80.1$ & \textcolor{darkcoral}{$81.4$} & $43.77$ & - & - & - \\ 
    SAC~\cite{Zhang_2017_ICCV}  &  ResNet-$101$  & $8\times$  & M & $78.1$ & - & $44.30$ & - & - & - \\
    AAF~\cite{aaf2018}   &   ResNet-$101$ & $8\times$ & R &  $79.1^{\flat}$  & - & - & - & - & - \\  
    DSSPN~\cite{Liang_2018_CVPR}  &  ResNet-$101$ & $8\times$ & - &  $77.8$ & - & $43.68$ & - & - & \textcolor{darkcoral}{$38.9$} \\ 
    DepthSeg~\cite{Kong_2018_CVPR} &  ResNet-$101$ & $8\times$  & - &  $78.2$ & - & - & - & - & - \\
    MMAN~\cite{luo2018macro} &  ResNet-$101$ & $8\times$  & - & - & - & - & \textcolor{blue}{$46.81$} & - & - \\
    JPPNet~\cite{liang2018look} &  ResNet-$101$ & $8\times$  & M & - & - & - & \textcolor{darkcoral}{$51.37$} & - & - \\
    EncNet~\cite{Zhang_2018_CVPR}   & ResNet-$101$ & $8\times$  & - & - & - & $44.65$ & - & {$51.7$} & - \\
    GCU~\cite{li2018beyond}   & ResNet-$101$  & $8\times$  & R & - & - & $44.81$ & - & - & -\\
    APCNet~\cite{he2019adaptive} & ResNet-$101$ & $8\times$ & M,R & - & - &  \textcolor{red}{$45.38$} & - & \textcolor{darkcoral}{$54.7$} & - \\
    CFNet~\cite{zhang2019co} & ResNet-$101$ & $8\times$ & R & $79.6$ & - & $44.89$ & - & \textcolor{blue}{$54.0$} & - \\
    BFP~\cite{ding2019boundary}   &  ResNet-$101$ & $8\times$ & R &  \textcolor{darkcoral}{$81.4$} & - & - & - & $53.6$ & - \\
    CCNet~\cite{Huang_2019_ICCV}  &  ResNet-$101$ & $8\times$ & R & \textcolor{darkcoral}{$81.4$} & - & $45.22$ & - & - & - \\
    ANNet~\cite{Zhu_2019_ICCV}  &  ResNet-$101$ & $8\times$ & M,R &  $81.3$ & - & \textcolor{blue}{$45.24$} & - & $52.8$ & - \\
    \rowcolor{Gray}
    OCR (Seg. transformer) &   ResNet-$101$ & $8\times$ & R &\textcolor{red}{${81.8}$} & \textcolor{red}{${82.4}$} & \textcolor{darkcoral}{${45.28}$} & \textcolor{red}{${55.60}$} & \textcolor{red}{${54.8}$} & \textcolor{red}{${39.5}$} \\
    \hline 
    \multicolumn{10}{c}{Advanced baselines} \\
    \hline
    DenseASPP~\cite{Yang_2018_CVPR}  &  DenseNet-$161$  & $8\times$ & M &  $80.6$ & - & - & - & - & - \\
    DANet~\cite{fu2018dual}  &  ResNet-$101$ + MG  & $8\times$ & R &  $81.5$ & - & $45.22$ & - & $52.6$ & $39.7$ \\
    DGCNet~\cite{zhang2019dual} &   ResNet-$101$ + MG  & $8\times$ & R  &  $82.0$ & - & - & - & \textcolor{blue}{$53.7$} & - \\
    EMANet~\cite{Li_2019_ICCV} &   ResNet-$101$ + MG & $8\times$ & R  &  - & - & - & - &  $53.1$ & \textcolor{blue}{$39.9$} \\
    SeENet~\cite{Pang_2019_ICCV}  &  ResNet-$101$ + ASPP  & $8\times$ & M &  $81.2$ & - & - & - & - & - \\
    SGR~\cite{NIPS2018_7456}   & ResNet-$101$ + ASPP  & $8\times$ & R & - & - & $44.32$ & - & $52.5$ & $39.1$ \\ 
    OCNet~\cite{yuan2018ocnet}  &  ResNet-$101$ + ASPP & $8\times$ & M,R &  $81.7$ & - & \textcolor{blue}{$45.45$} & $54.72$ & - & - \\
    ACFNet~\cite{Zhang_2019_ICCV} &  ResNet-$101$ + ASPP & $8\times$ & M,R &  $81.8$ & - & - & - & - & - \\
    CNIF~\cite{Wang_2019_ICCV} &   ResNet-$101$ + ASPP  & $8\times$ & M &  - & - & - & \textcolor{red}{$56.93$} & - \\
    GALD~\cite{xiangtl_gald} &  ResNet-$101$ + ASPP & $8\times$ & M,R &  $81.8$ & $82.9$ & - & - & - & -\\
    GALD{$^\dag$}~\cite{xiangtl_gald} &  ResNet-$101$ + CGNL + MG & $8\times$ & M,R &  - & \textcolor{blue}{${83.3}$} & - & - & - & -\\
    Mapillary~\cite{Bulo_2018_CVPR} &  WideResNet-$38$ + ASPP  & $8\times$ & M & - &  $82.0$ & - & - & - & - \\
    GSCNN{$^\dag$}~\cite{gscnn} &  WideResNet-$38$ + ASPP  & $8\times$ & M &  $\textcolor{darkcoral}{82.8}$ & - & - & - & - \\ 
    SPGNet~\cite{Cheng_2019_ICCV} &  $2\times$ ResNet-$50$ & $4\times$ & - &  $81.1$ & - & - & - & - & - \\ 
    ZigZagNet~\cite{Lin_2019_CVPR} & ResNet-$101$ & $4\times$ & M & - & - & - & - & $52.1$ & - \\
    SVCNet~\cite{ding2019semantic}  & ResNet-$101$ & $4\times$ & R & $81.0$ & - & - & - & $53.2$ & $39.6$\\
    ACNet~\cite{Fu_2019_ICCV} &  ResNet-$101$ + MG  & $4\times$ & M,R &  \textcolor{blue}{$82.3$} & - & \textcolor{red}{${45.90}$} & - & \textcolor{darkcoral}{$54.1$} & \textcolor{darkcoral}{$40.1$} \\
    CE2P~\cite{liu2018devil}   &   ResNet-$101$ + PPM & $4\times$ & M &  - & - & - & $53.10$ & - \\
    VPLR{$^{\dag\ddag}$}~\cite{zhu2019improving} &  WideResNet-$38$ + ASPP  & $4\times$ & M &  - & $\textcolor{darkcoral}{83.5}$ & - & - & - \\ 
    DeepLabv3+~\cite{chen2018encoder} &  Xception-$71$ & $4\times$ & M & - &  $82.1$ & - & - & - & - \\
    DPC~\cite{chen2018searching} & Xception-$71$  & $4\times$ & M &  ${82.7}$  & - & - & - & - \\
    DUpsampling~\cite{Tian_2019_CVPR} & Xception-$71$ & $4\times$ & M & - & - & - & - & $52.5$ & - \\
    HRNet~\cite{sun2019high} &   HRNetV2-W$48$  & $4\times$ & - &  $81.6$ & - & - & \textcolor{blue}{$55.90$} & $54.0$ \\
    \rowcolor{Gray}
    OCR (Seg. transformer) &   HRNetV2-W$48$ & $4\times$ & R & $82.4$ & ${83.0}$ & \textcolor{darkcoral}{$45.66$} & \textcolor{darkcoral}{$56.65$} & \textcolor{red}{$56.2$} & \textcolor{red}{$40.5$} \\
    \rowcolor{Gray}
    OCR{$^\dag$} (Seg. transformer) &   HRNetV2-W$48$  & $4\times$ & R & \textcolor{red}{$83.6$} & \textcolor{red}{$84.2$} & - & - & - & - \\
    \shline
    \end{tabular}
}
\label{table:ocnet_sota_exp_all}
\end{table*}

For comparison with the approaches based on the advanced baselines,
we perform our OCR on the HRNet-W48, and pre-train our model on the Mapillary dataset~\cite{neuhold2017mapillary}.
Our approach achieves $84.2\%$ on Cityscapes \texttt{test}.
\textcolor{black}
{
We further apply a novel post-processing scheme SegFix~\cite{yuan2020segfix}
to refine the boundary quality, which brings $0.3\%\uparrow$ improvement.
Our final submission ``HRNet + OCR + SegFix" achieves $84.5\%$, which ranks the \nth{1} place on the Cityscapes leaderboard by the time of our submission.
}
In fact, we perform PPM and ASPP on HRNet-W$48$ separately 
and empirically find that directly applying either PPM or ASPP 
does not improve the performance
and even degrades the performance,
while our OCR consistently improves the performance.

Notably, the very recent work~\cite{tao2020hierarchical} sets a new state-of-the-art performance $85.4\%$ on Cityscapes leaderboard via combining our ``HRNet + OCR'' and a new hierarchical multi-scale attention mechanism.

\noindent\textbf{ADE20K.}
From Table~\ref{table:ocnet_sota_exp_all},
it can be seen that
our OCR achieves competitive performance ($45.28\%$ and $45.66\%$) compared with most of the previous approaches
based on both simple baselines and advanced baselines.
For example, the ACFNet~\cite{he2019adaptive} exploits both the multi-scale context and relational context to achieve higher performance.
The very recent ACNet~\cite{Fu_2019_ICCV} achieves the best performance through combining richer local and global contexts.

\noindent\textbf{LIP.}
Our approach achieves the best performance 
$55.60\%$ on LIP \texttt{val} based on the simple baselines.
Applying the stronger backbone HRNetV2-W$48$ further improves the performance
to $56.65\%$, 
which outperforms the previous approaches.
The very recent work CNIF~\cite{Wang_2019_ICCV} achieves the best performance ($56.93\%$)
through injecting the hierarchical structure knowledge of human parts.
Our approach potentially benefit from such hierarchical structural knowledge.
All the results are based on only flip testing without multi-scale testing\footnote{Only few methods adopt multi-scale testing. For example, 
CNIF~\cite{Wang_2019_ICCV}
gets the improved performance
from $56.93\%$ to $57.74\%$.}.

\noindent\textbf{PASCAL-Context.}
We evaluate the performance over $59$ categories following~\cite{sun2019high}.
It can be seen that 
our approach outperforms both the previous best methods based on simple baselines and the previous best methods based on advanced baselines.
The HRNet-W$48$ + OCR approach achieves the best performance $56.2\%$, 
significantly
outperforming the second best, e.g., ACPNet ($54.7\%$) and ACNet ($54.1\%$).

\noindent\textbf{COCO-Stuff.}
It can be seen that 
our approach achieves the best performance, $39.5\%$ based ResNet-$101$ and $40.5\%$ based on HRNetV2-$48$.

\noindent\textbf{Qualitative Results.}
We illustrate the qualitative results 
in the supplementary material due to the limited pages.

\section{Experiments: Panoptic Segmentation}

To verify the generalization ability of our method,
we apply OCR scheme on the more challenging panoptic segmentation task~\cite{kirillov2019panopticseg}, 
which unifies both the instance segmentation task and
the semantic segmentation task.

\noindent\textbf{Dataset.}
We choose the COCO dataset~\cite{lin2014microsoft} to study the effectiveness of our method on panoptic segmentation.
We follow the previous work~\cite{kirillov2019panoptic} and uses all 2017
COCO images with 80 thing and 53 stuff classes annotated.

\begin{table*}[t!]
\centering
\caption{
\textbf{Panoptic segmentation results on COCO val 2017.} The performance of Panoptic-FPN~\cite{kirillov2019panoptic} is reproduced based on the official
open-source Detectron2~\cite{wu2019detectron2}
and we use the $3\times$ learning rate schedule by default. Our OCR consistently improves the PQ performance with both backbones.}
\small
\resizebox{0.9\linewidth}{!}
{
\begin{tabular}{c|l|p{1.8cm}p{1.8cm}|p{1.8cm}p{1.8cm}|p{1.8cm}}
\shline
 Backbone & Method & AP & PQ\textsuperscript{Th} & mIoU & PQ\textsuperscript{St} & PQ \\
\shline
\multirow{2}{*}{ResNet-50} & Panoptic-FPN & 40.0 & 48.3 & 42.9 & 31.2 & 41.5 \\
& Panoptic-FPN + OCR  & {40.4} (+\emph{0.4}) & {48.6} (+\emph{0.3}) & {44.3} (+\emph{1.4}) & {33.9} (+\emph{2.7}) & {42.7} (+\emph{1.2}) \\
\hline
\multirow{2}{*}{ResNet-101} & Panoptic-FPN & 42.4 & 49.7 & 44.5 & 32.9 & 43.0 \\
& Panoptic-FPN + OCR  & {42.7} (+\emph{0.3}) & {50.2} (+\emph{0.5}) & {45.5} (+\emph{1.0}) & {35.2} (+\emph{2.3}) & {44.2} (+\emph{1.2}) \\
\shline
\end{tabular}
\label{table:panoptic}
}
\end{table*}

\noindent\textbf{Training Details.}
We follow the default training setup of ``COCO Panoptic Segmentation Baselines with Panoptic FPN ($3\times$ learning schedule)''
\footnote{\rm{https://github.com/facebookresearch/detectron2/blob/master/MODEL\_ZOO.md}} in Detectron2~\cite{wu2019detectron2}.
The reproduced Panoptic FPN reaches higher performance than the original numbers in the paper~\cite{kirillov2019panoptic} (Panoptic FPN w/ ResNet-50, PQ: $39.2\%$ / Panoptic FPN w/ ResNet-101, PQ: $40.3\%$) and we choose the higher reproduced results as our baseline.

In our implementation,
we use the original prediction from the semantic segmentation head (within Panoptic-FPN) to compute the soft object regions and then we use a OCR head to predict a refined
semantic segmentation map.
We set the loss weights on both the original semantic segmentation head and the OCR head as $0.25$.
All the other training settings are kept the same for fair comparison.
We directly use the same OCR implementation (for the semantic segmentation task) without any tuning.

\noindent\textbf{Results.}
In Table~\ref{table:panoptic},
we can see that OCR improves the PQ performance of Panoptic-FPN (ResNet-$101$) from $43.0\%$ to $44.2\%$, where the main improvements
come from better segmentation quality on the \emph{stuff} region measured by
mIoU and PQ\textsuperscript{St}.
Specifically,
our OCR improves the mIoU and PQ\textsuperscript{St} of
Panoptic-FPN (ResNet-$101$)
by $1.0\%$ and $2.3\%$ separately.
In general, the performance of ``Panoptic-FPN + OCR" is very competitive compared to various recent methods~\cite{xiong19upsnet,liu2019end,yang2019sognet}.
We also report the results of Panoptic-FPN with PPM and ASPP to
illustrate the advantages of our OCR in the supplementary material.

\section{Conclusions}
In this work,
we present an object-contextual representation
approach for 
semantic segmentation.
The main reason for the success  
is that the label of a pixel
is the label of the object
that the pixel lies in
and the pixel representation
is strengthened
by characterizing each pixel
with the corresponding object region representation.
We empirically show that our approach
brings consistent improvements on various benchmarks.

\noindent \textbf{Acknowledgement}
This work is partially supported by Natural Science Foundation of China under contract No. 61390511, and Frontier Science Key Research Project CAS No. QYZDJ-SSW-JSC009.

\bibliographystyle{splncs04}
\small
\bibliography{ocr}
\end{document}